\documentclass[journal]{IEEEtran}
\IEEEoverridecommandlockouts
\usepackage{cite}
\usepackage{amsmath,amssymb,amsfonts}
\usepackage{algorithmic}
\usepackage{graphicx}
\usepackage{textcomp}
\usepackage{xcolor}
\usepackage{amssymb}% http://ctan.org/pkg/amssymb
\usepackage{pifont}% http://ctan.org/pkg/pifont
\newcommand{\xmark}{\ding{55}}%
\def\BibTeX{{\rm B\kern-.05em{\sc i\kern-.025em b}\kern-.08em
    T\kern-.1667em\lower.7ex\hbox{E}\kern-.125emX}}
\begin{document}

\title{Deep Learning for Visual Navigation of Underwater Robots\\
}

\author{\IEEEauthorblockN{MD Sunbeam} \\
\IEEEauthorblockA{\textit{Department of Aerospace Engineering} \\
\textit{Texas A\&M University}\\
College Station, TX, United States \\
mdsunbeam@tamu.edu}
}

\maketitle

\begin{abstract}
This paper aims to briefly survey deep learning methods for visual navigation of underwater robotics. The scope of this paper includes the visual perception of underwater robotics with deep learning methods, the available visual underwater datasets, imitation learning, and reinforcement learning methods for navigation. Additionally, relevant works will be categorized under the imitation learning or deep learning paradigm for underwater robots for clarity of the training methodologies in the current landscape. Literature that uses deep learning algorithms to process non-visual data for underwater navigation will not be considered, except as contrasting examples.
\end{abstract}


\section{Introduction}

This paper will cover deep learning for underwater robotics. It is divided into the perception of underwater robotics, underwater visual datasets, imitation learning, and reinforcement learning for underwater environments.

Visual navigation in underwater robotics is more challenging than its land or air counterparts because of the limitations of perception, particularly computer vision. When light travels through the water, it is absorbed and scattered, resulting in a wavelength-dependent attenuation that disturbs the standard ways of handling vision \cite{a1}. An example of this problem is capturing images of an object with different brightness and colors based on distance. 

% \begin{figure}[htbp]
% \centerline{\includegraphics[scale=.35]{cv_challenges.png}}
% \caption{Figure taken from \cite{a1}. (a) shows low visibility caused by high scattering, (b) shows how visible colors changed from their original underwater due to absorption, (c) show the heterogeneous illumination of the seabed through a flashlight, and (d) shows floating particles blocking the sand floor}
% \label{fig}
% \end{figure}

Additionally, environmental changes are more subtle underwater. Features may change from drifting sand and turbulent current. Detecting and modeling these changes is difficult, especially because of its stochastic nature and the complex fluid equations that govern the dynamics. The hydrodynamics involved is nonlinear and time-varying, and the robot is usually underactuated with not all the states reachable \cite{a2}. Deep learning methods for visual navigation are explored, because they offer a learning-based way that may handle these challenges through a data-driven approach. 

Because autonomous underwater vehicles (AUVs) are less common due to greater hardware considerations, underwater datasets are rare and more difficult to collect. Since deep learning is a data-driven method, it is imperative to have public datasets to benchmark models.  

The two main deep learning paradigms for control of an AUV through visual navigation is imitation and reinforcement learning. Imitation learning is used for underwater robotics for visual navigation, often mapping RGB images to control commands like roll, pitch, yaw, and throttle. The problem formulation is a supervised learning task where the objective is to learn a policy that imitates the trajectories of a human or robot demonstrator. Two major limitations of imitation learning algorithms are compounding error between differences of expert and agent trajectory as well as the idea that a policy derived from such a method will not outperform the expert demonstrator \cite{a4}.

Reinforcement learning is the other deep learning paradigm, using a non-differentiable objective function through rewards to train neural networks for a desired task, in this case being underwater visual navigation. Reinforcement learning differs from imitation learning in that an initial dataset is not needed, and an environment must be explored to generate the data to be trained on. Though reinforcement learning is less sample efficient than imitation learning, the lack of dependence on an expert demonstrator allows it to yield policies that are potentially superhuman \cite{a15}.

The paper's contribution are as follows:

\begin{itemize}
\item Identify the works that focus on utilizing deep learning for perception in underwater visual navigation algorithms.
\item Discuss the publicly available underwater datasets.
\item Categorize the deep learning methods used in AUVs under imitation learning or reinforcement learning.
\end{itemize}

%\section{Ease of Use}

\section{Perception}

Due to the challenges of visual perception in underwater robotics, \cite{a10} offers a deep learning method to improve loop closure and detection for a Visual Graph SLAM algorithm for underwater navigation. Siamese networks, a neural network architecture that has enjoyed great success in similarity learning, was leveraged to detect similar visual features and places. 

Another work for underwater loop detection for use in a visual SLAM algorithm is \cite{a11}, but it sets itself apart from \cite{a10} by training a network through an unsupervised learning method. Whereas Siamese networks are trained in a supervised manner with image pairs through contrastive loss, the unsupervised learning method detects similar images and loop closure through the clustering of image descriptors.

\begin{figure}[htbp]
\centerline{\includegraphics[scale=.3]{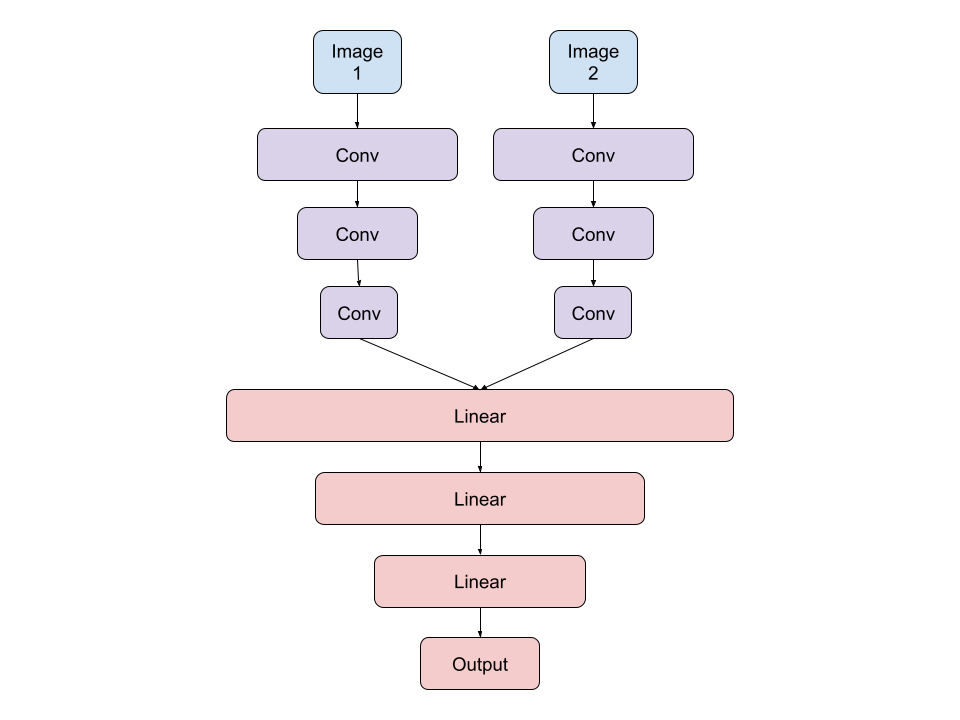}}
\caption{Redrawn network architecture diagram of the Siamese network used to detect loop-closure between two underwater images in \cite{a10}.}
\label{fig}
\end{figure}

Finally, \cite{a9} also deals with loop closure, but the role of the neural network is more limited in scope. The neural network only selects loop candidates, which are sent to an image matcher and geometric consistency verifier to output the final loop detected images.

\section{Underwater Datasets}

There are far fewer underwater robot datasets as compared to autonomous ground vehicle or autonomous aerial vehicle datasets since collecting data underwater remains difficult due to hardware constraints and needing a relatively uncommon environment. 

For deep learning methods, having a wide variety of datasets is paramount because it allows the research community to tackle the same problems, often on a vetted or curated dataset. In computer vision, deep learning innovation was primarily accelerated by the existence of ImageNet \cite{a19}. In natural language processing, the large corpus of textual data in the internet has fueled the advances of large language models. For robotics, ground vehicle datasets like CARLA \cite{a20} or KITTI \cite{a21} and aerial vehicle datasets like Mid-Air \cite{a22} have paved the way for development of many deep learning based visual navigation algorithms.

As such, \cite{a6} offers a visual dataset for AUVs navigating near the seabed. The images were collected from three different depths: one at a few meters, the second 270 meters, and the last at 380 meters. The purpose of this data is to provide image and inertia-pressure samples for use in simultaneous localization and mapping algorithms.    

% \begin{figure}[htbp]
% \centerline{\includegraphics[scale=.32]{aqualoc_dataset.png}}
% \caption{These are some samples images from the AQUALOC dataset in \cite{a6}}
% \label{fig}
% \end{figure}

A common theme is leveraging generative adversarial networks and generative models to augment real underwater datasets with generated synthetic images. This is done in \cite{a7}, \cite{a17}, and \cite{a18}. A publicly available dataset, called EUVP, of images that were taken during ocean experiments for exploration and human-robot collaboration is also presented in \cite{a18}. 

\begin{figure}[htbp]
\centerline{\includegraphics[scale=.47]{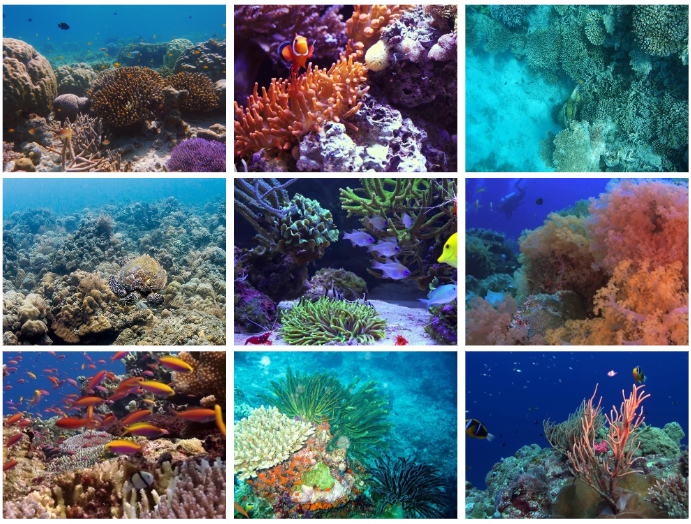}}
\caption{These are some sample training images selected from the EUVP dataset in \cite{a18}.}
\label{fig}
\end{figure}

\section{Imitation Learning}

For \cite{a3}, a domain expert diver collected data to generate "good" and "bad" navigation scenarios, which were later annotated with labels of yaw and pitch to accomplish the task of exploring a coral reef while avoiding obstacles. A convolutional neural network was trained to map the images to the control commands. They evaluated their models through the percentage of coral covered in the navigation task in certain reef area.

% \begin{figure}[htbp]
% \centerline{\includegraphics[scale=.45]{annotation_tool.png}}
% \caption{This is the annotation interface for the demonstration dataset in \cite{a3}}
% \label{fig}
% \end{figure}

Similar to \cite{a3}, \cite{a4} also uses a behavior cloning model to map RGB images to yaw and pitch command. In this case, the task was to explore a shipwreck, and a convolutional neural network was used. For this work, the neural network was trained on a mixture of simulation and real-world data. The model was evaluated through the training-validation-test split. Different test accuracies are given for the real-world only dataset, simulation only dataset, and mixed dataset. The primary difference between the two papers were the difference in task and the nature of the demonstration data.

One work that is different is \cite{a5}, which uses goal-conditioned imitation learning for underwater visual navigation. The behavior cloning model learns safe, reactive behavior for difficult terrain, and is conditioned to navigate based on waypoints. The neural network architecture is a convolutional neural network, much like those in \cite{a3} and \cite{a4}. Similarly, the model was evaluated through a test set and evaluated in real-life qualitatively. 

%\cite{a3}
%\cite{a4}
%\cite{a5}

\section{Reinforcement Learning}

In \cite{a8}, a soft-actor critic deep reinforcement learning algorithm was used to train a neural network. The AUV was a soft robot, and the task was to swim in a straight line in disturbed water. A camera was used to collect RGB images, which served as the observation space for the neural network. Before the deep reinforcement learning algorithm was rolled out underwater, a model was trained in a MuJoCo simulation.

\begin{figure}[htbp]
\centerline{\includegraphics[scale=.33]{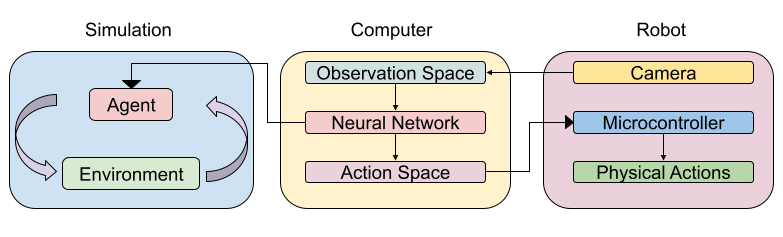}}
\caption{Redrawn system diagram for the deep reinforcement learning controller in \cite{a8}.}
\label{fig}
\end{figure}

For \cite{a15}, a combination of imitation learning and reinforcement learning is used. First, a generative adversarial imitation learning algorithm is used to overcome the cold start problem of the initial neural network training to learn a policy. Then, a reward function is designed and trained with proximal policy optimization and soft-actor critic. The results are compared in a Unity simulation. This work uses a light sensor, which is different from the other imitation learning and reinforcement learning work as most use a RGB camera for their visual sensor.

\section{Imitation and Reinforcement Learning Categorization}

The works where deep learning methods were used for only visual navigation are considered and categorized. Works like \cite{a12}, \cite{a14}, and \cite{a16} use deep or reinforcement learning methodologies, but fail to incorporate a visual sensor. While there is much more work on using neural networks on inertial, pressure, and position data from non-visual sensors, that is outside the scope of this survey. Furthermore, the specific neural network architectures used in underwater robot navigation will not be discussed in depth like in \cite{a13}, but are useful for understanding their applications in perception and control.
\vspace{4mm}

\begin{tabular}{ |p{2cm}||p{2cm}||p{2cm}| }
 \hline
 \multicolumn{3}{|c|}{Deep Learning Paradigm Categorization} \\
 \hline
 Reference & Imitation Learning & Reinforcement Learning\\
 \hline
 \cite{a3} & \checkmark & \xmark\\
 \cite{a4} & \checkmark & \xmark\\
 \cite{a5} & \checkmark & \xmark\\
 \cite{a8} & \xmark & \checkmark\\
 \cite{a15} & \checkmark & \checkmark\\
 \hline
\end{tabular}
\vspace{4mm}

There are far more works using imitation learning algorithms than reinforcement learning. This can be explained by the idea that the simplest class of imitation learning algorithms, behavior cloning, is good at dealing with high dimensional inputs because convolutional layers can be used as a feature extractor to reduce the dimension of the images. The behavior cloning problem formulation is also easier, where the objective is imitating an expert trajectory through a supervised learning setup.

The lack of many deep reinforcement learning works for visual navigation of AUVs can be explained by the added difficulty of devising a proper reward function on top of dealing with a noisy observation space. Moreover, the reinforcement learning agent must do exploration before it can exploit through an adequate policy, which introduces an overhead of creating or using a simulation for the underwater robot as real-life exploration can be expensive and dangerous. 

\section{Conclusion}

In this paper, we have covered the deep learning methods used to improve perception under water, which go into improving the visual navigation algorithms that use those modules. Most of the work centers around in detecting loop closures through similarity learning of images through Siamese networks or using other neural network architectures to detect and select loop closure candidates like in \cite{a9}, \cite{a10}, and \cite{a11}. The improvement of loop closure in visual perception through deep learning improves the visual SLAM algorithms that leverage them for navigation. 

Next, we discussed the underwater datasets publicly available like the ones in \cite{a6} and \cite{a7}. While there are some standard public datasets, there are far too few and even less for a robotics application. This can be attributed to the difficulty of creating such datasets due to hardware and environmental constraints. A common trend is to leverage deep learning methods to augment the available datasets, either through improving the quality of the images or generating synthetic images like in \cite{a17} and \cite{a18}. However, this type of augmentation seems more a stopgap rather than a permanent solution to the scarcity of underwater robotics data. 

Finally, we categorize the works that use deep learning methods to control an AUV under the imitation (like in \cite{a3}, \cite{a4}, and \cite{a5}) or reinforcement learning (like in \cite{a8} and \cite{a15}) paradigm. A pattern that becomes immediately obvious is that there exists far more imitation learning works than deep reinforcement learning works in the domain of visual navigation in underwater robotics. Another trend is that there aren't that many works where some visual sensors are used in the context of imitation or reinforcement learning. This can perhaps be explained by the hardware constraints in AUVs, where limited computation and batteries onboard may make RGB cameras too energy-intensive. 

From this survey, some gaps in the field of deep learning for visual navigation of underwater robots become apparent. There needs to be more underwater datasets collected and made publicly available, especially with robotics applications. Even though deep learning methods are used to augment existing underwater datasets, the best way to accelerate the pace and scalability of neural networks is through more data. Like in computer vision and natural language processing, curated datasets provide a standard for research and competition, both of which push innovation within the field.    

Another gap is the lack of reinforcement learning work for visual navigation. The preference for imitation learning is reasonable given the harder formulation of the reinforcement learning problem, but focus in this direction may address the limitations of imitation learning, like compounding errors and the learned policy never being better than the demonstrator. Because simulation is an important component of reinforcement learning in the exploration stage, attention in this area may lead to better simulators that can address the lack of underwater datasets through synthetic data.

Overall, the field of deep learning for visual navigation of underwater robots provides natural challenges crucial for the larger study of AI robotics. One such challenge is the fundamental problem of acting under noisy or misleading visual perception data. This offers opportunities for planning or navigation under uncertainty. Moreover, breakthroughs in the domain of deep learning for underwater visual navigation may be applied to the broader field of learning from sparse environments. Underwater environments are often characterized by sparsely distributed features. In imitation learning and deep reinforcement learning, environments with few features may make it difficult for the neural network to learn a useful policy for the desired task. Because of these reasons, it is crucial to place more research emphasis on the problem of visual navigation in underwater environments.

\vspace{12pt}
\end{document}